\documentclass{article} %
\usepackage{iclr2027_conference,times}

\usepackage{amsmath,amsfonts,bm}

\def\eqref#1{equation~\ref{#1}}

\def\1{\bm{1}}

\DeclareMathAlphabet{\mathsfit}{\encodingdefault}{\sfdefault}{m}{sl}
\SetMathAlphabet{\mathsfit}{bold}{\encodingdefault}{\sfdefault}{bx}{n}

\newcommand{\R}{\mathbb{R}}

\usepackage{hyperref}
\usepackage{url}

\title{Autoregressive Frontier Expansion: \\Growing Trees with Graph Machine Learning}

\author{Umer Gupta \\
Independent Researcher\\
London, United Kingdom \\
{\fontsize{8}{9}\selectfont\texttt{umer.gupta152@gmail.com}} \\
\And
Saku Peltonen \\
ETH Zurich \\
Zurich, Switzerland \\
{\fontsize{8}{9}\selectfont\texttt{speltonen@ethz.ch}} \\
\And
Martin Ritzert \\
Leipzig University \\
Leipzig, Germany \\
{\fontsize{8}{9}\selectfont\texttt{martin.ritzert@uni-leipzig.de}}
}

\newcommand{\metricdef}[1]{%
  \par\addvspace{.25\baselineskip}%
  \noindent\emph{#1}:\enspace\ignorespaces
}

\newif\ifdraft
\draftfalse

\usepackage{tikz}
\usetikzlibrary{arrows.meta, positioning, calc, shapes.geometric}

\usetikzlibrary{calc, arrows.meta, backgrounds}
\definecolor{cbrown}{RGB}{93,73,64}
\definecolor{corange}{RGB}{255,167,123}
\colorlet{tgBaseFill}{cbrown!10}
\colorlet{tgBaseDraw}{cbrown!20!black}
\tikzset{
    tg/edge/.style={line width=1pt, draw=black!75, line cap=round},
    tg/node/.style={circle, draw=tgBaseDraw, thick, fill=tgBaseFill, minimum size=4mm, inner sep=0pt, font=\scriptsize},
}

\usepackage{multirow}
\usepackage{makecell}
\usepackage[ruled,vlined,noend,linesnumbered]{algorithm2e}
\usepackage{amsmath}
\usepackage[capitalize,]{cleveref}
\usepackage{booktabs}
\usepackage{placeins}

\iclrpreprintcopy
\begin{document}

\maketitle

\begin{abstract}
    Tree-like branching structures are common in nature, from botanical trees to neurons, blood vessels and respiratory trees. Their branching shape often reflects function, making structural modelling central to understanding how these systems work. Because acquiring real-world 3D data is often expensive or infeasible, realistic generative models are valuable for simulation and data augmentation. Existing morphology-specific models either constrain how topology is generated or rely on hand-tuned, mechanistic procedures. Generic 3D graph generators, by contrast, do not exploit or enforce the structure of trees. We propose Autoregressive Frontier Expansion, a generative framework that constructs trees through an iterative expansion process, simulating the biological growth of real trees. At each step, a flow-matching model parameterised by an SO(2)-equivariant GNN expands the frontier by predicting whether each active branch bifurcates or terminates. We evaluate our method on cortical neurons and botanical trees in unconditional, class-conditioned, and morphology-guided generation. Across both domains, the generated morphologies agree closely with the reference distributions and, in conditional experiments, with the specified targets.
\end{abstract}
    
\section{Introduction}

    Tree-like (acyclic) branching structures regularly appear in nature: respiratory trees branch to increase surface area for gas exchange; botanical trees branch to improve light capture and resource distribution; and neurons branch to form connections with other neurons. Across biological systems, morphology -- the anatomical shape -- is often closely tied to physical constraints and functional requirements. For example, neuronal morphologies define the wiring patterns of the brain's circuitry and consequently its function \citep{torben2014context, memelli2013self, lin2018evolutionary}.
    Generating realistic simulations of branching structures is, hence, valuable for understanding such biological systems, especially where data capture is difficult. 
    The brain, for example, contains billions of neurons with diverse morphologies \citep{markram2004interneurons} that need to be mapped to understand its function. This is resource-intensive \citep{schmitz2011automated}, even for volumes as small as a cubic millimetre of a mammalian brain \citep{microns2025functional}. 
    Beyond neuroscience, inpainting incomplete scans of trees directly leads to better estimation of their biomass.
    In both domains, branching structure is extracted from point clouds and data augmentation with realistic synthetic structures can improve this step. 
    
    Several classical approaches have been used to generate branching morphologies \citep{honda1971description, prusinkiewicz1990graphical, kanari2022computational}, but work on learning-based methods remains limited. 
    MorphGrower \citep{yang2024morphgrower} generates neuronal morphologies layer-by-layer, mimicking the growth pattern in nature, but uses a \emph{predefined} branching structure from a seed tree -- only the geometry, not the structure, is predicted. 
    Similarly, MorphoGen \citep{zhu2025morphogen} predicts geometry through a point cloud while the tree is recovered post hoc by a skeletonisation heuristic.
    In both cases the topology is not explicitly learned but either fixed in advance or recovered afterwards.
    
    In this work, we propose \emph{autoregressive frontier expansion}, an iterative procedure that conditions future branching on the intermediate morphology, mirroring how branching structures grow in nature \citep{barthelemy2007plant, niell2004vivo}. In contrast to previous approaches, our method predicts both \emph{topology} and \emph{geometry} jointly. We parameterise this process with an SO(2)-equivariant GNN that accounts for the principal axis of biological structures. Our method is able to generate structurally valid trees by construction. Across $26{,}000$ cortical neurons as well as botanical trees, the generated morphologies closely match the corresponding reference distributions. Further, our method follows a target's branching profile when guided by it, making it well suited for data augmentation. Lastly, iterative generation is computationally efficient and scales better with tree size compared to baselines.
    
    \begin{figure}[t]
        \centering
        \scalebox{0.92}{\input{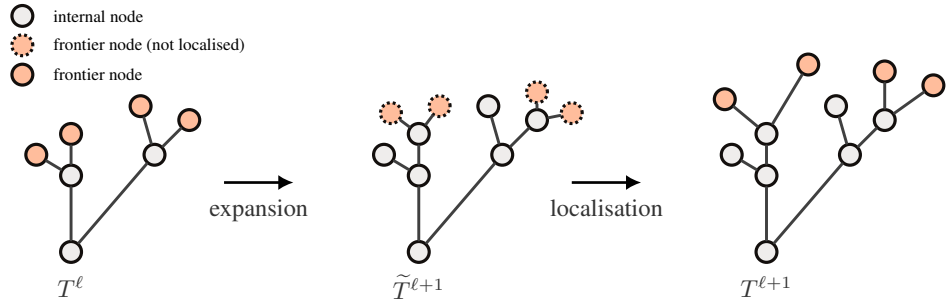}}
        \caption{Overview of a generation step. Starting from a partial tree $T^\ell$, we (i) \textbf{expand} the current frontier (a subset of leaves) by deciding for each frontier node whether it branches or terminates, yielding a tree $\widetilde{T}^{\ell + 1}$; then (ii) \textbf{localise} by predicting 3D parent-relative offsets for the previous frontier nodes, producing $T^{\ell+1}$. The process repeats until all frontier nodes terminate.}
        \label{fig:iterativeprocedure}
    \end{figure}
    
    \subsection{Related Work}
    \label{section: related work}
    
    \paragraph{Growth-rule based approaches} On the classical, non-learning side, growth-rule based methods use a predefined set of rules or probabilistic conditioning on priors to model the growth process of branching morphologies \citep{yang2024morphgrower}. 
    \citet{honda1971description} uses preset rules to model a tree as repeated bifurcations of straight segments with fixed branching angles and constant length contraction ratios. 
    Topological Neuron Synthesis (TNS) \citep{kanari2022computational} simulates growth as a probabilistic procedure that generates neuronal trees conditioned on a topological description of their branching structure. 
    The conditioning signal is given by persistence barcodes \citep{carlsson2009topology}, formalised for complex branching morphologies by \citet{kanari2018topological} as the Topological Morphology Descriptor (TMD) which we also use in our experiments.
    \citet{li2021learning} combines L-systems with learning and conditions on individual images.
    
    \paragraph{Learned tree generation} Existing learned methods for generating neuronal trees include MorphVAE \citep{laturnus2021morphvae}, which trains a sequence-to-sequence variational autoencoder on soma-to-tip walks. New trees are sampled by decoding a set of such walks in 3D, clustering nearby points to merge nodes, and connecting consecutive points along each walk to form edges. Because the walks are decoded independently and merged only post hoc, MorphVAE does not guarantee a globally consistent tree topology.
    MorphGrower \citep{yang2024morphgrower} grows neuronal trees iteratively, similar to our method.
    However, the branching structure is fixed to that of a seed tree, and only new geometry is generated. 
    Thus, neither method performs the end-to-end joint generation of topology and geometry considered in our experiments.
    MorphoGen \citep{zhu2025morphogen} instead generates the complete point cloud before reconstructing the tree topology. A DiT-3D diffusion model \citep{mo2023dit} generates the neuron as an unordered cloud of $2{,}048$ points. A tree is recovered afterwards by L1-medial skeletonisation and a greedy linking heuristic. 
    The connections are therefore added after generation rather than predicted by the model.
    We evaluate MorphoGen in the unconditional setting; conditional generation is not supported by the model.
    \citet{lee2023latent} directly produce L-system-like strings using a transformer but relies on a large synthetic training set instead of real-world trees.    
    
    More broadly, TreeGen \citep{kollovieh2025treegen}  infers a hierarchical structure based on leaf positions in high-energy physics, thus focusing on the topology.
    Topology-only tree generation further appears as a subroutine in general learnable graph generation algorithms. For example, \citet{jin2018junction} generate molecules by generating and expanding the tree backbone of the molecule and \citet{shirzad2022td} expand the idea to general graphs. 
    \citet{bergmeister2023efficient} use iterative expansion and diffusion for generation of graphs without node features.
    
    \section{Methodology}
    
    A tree $T$ is a tuple $(V, E, P)$ where $V=\{1,\dots,N\}$ is the node set, $E\subset V\times V$ is the edge set,
    and $P\in\mathbb{R}^{|V|\times 3}$ stores the node coordinates (root-centred). We write $r\in V$ for the root. Each non-root node $v \neq r$ has a unique parent $\pi(v)$. Let $\operatorname{Leaves}(T)\subseteq V$ denote the set of leaves of the tree. For any node-indexed matrix $M$ we write $M(v)$ for the row of node $v$.

    For modelling trees we restrict ourselves to the overarching branching geometry and thus only consider branching points (bifurcations) and leaves (terminations/tips), see Appendix \ref{app:preprocessing}. 
    We can do so since most information is encoded in the `orientation and length of new branches' for both botanical trees \citep{honda1971description} and neurons \citep{kanari2018topological}.
    
    \subsection{Tree Expansion}
    
    We generate trees iteratively as a sequence $\{T^0, ..., T^L\}$. A partial tree at level $\ell$ is $T^\ell=(V^\ell,E^\ell,P^\ell)$. Our generation operates on an \emph{active frontier} $\mathcal{A}^\ell = V^\ell \setminus V^{\ell-1}\subseteq \operatorname{Leaves}(T^\ell)$, i.e., the leaves created in the previous step. Each iteration consists of two operations: (1) frontier expansion and (2) localisation. See \cref{fig:iterativeprocedure} for an overview and Appendix \ref{app:pseudocode} for pseudocode.

    \paragraph{Frontier expansion.}
    Given $T^\ell$, we predict a binary expansion state $\Gamma^{\ell}(v)\in\{0,1\}$ for each $v\in\mathcal{A}^\ell$ and attach two children $(v_L,v_R)$ to every $v$ with $\Gamma^{\ell}(v) = 1$.
    Together with the pairs of new edges $(v,v_L)$ and $(v,v_R)$ this defines the intermediary expanded tree $\tilde T^{\ell+1}$.
    
    \paragraph{Localisation.}
    In the intermediary tree $\tilde T^{\ell+1}$ the coordinates of the new leaves $v \in \mathcal{A}^{\ell+1}$ are not yet specified.
    For each such $v\in \mathcal A^{\ell+1}$ we get the coordinates $P^{\ell+1}(v)$ and thus the next partial tree $T^{\ell+1}$ by predicting parent-relative offsets $C^{\ell+1}(v)$:
    \begin{align}
        P^{\ell+1}(v) = P^\ell\big({\pi(v)}\big) + C^{\ell+1}(v),
        \label{eq:relative_position_parametrisation}
    \end{align}
    Offsets $C^{\ell+1}(v)$ are predicted conditioned on the intermediary tree $\tilde T^{\ell+1}$.
    For all nodes not in the frontier $\mathcal A^{\ell+1}$, the position does not change, so $P^{\ell+1}(u)=P^{\ell}(u)$ for $u\notin \mathcal A^{\ell+1}$. 

    \begin{figure}
        \centering
        \includegraphics[width=0.97\linewidth]{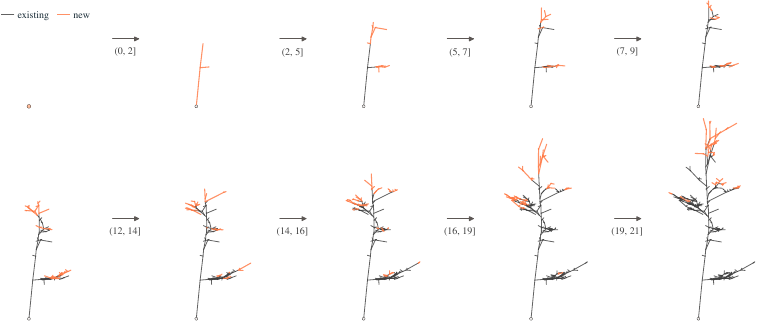}
        \caption{Stepwise generation of a botanical tree}
        \label{fig:stepwise}
    \end{figure}
    
    \subsection{Modelling and Training}
    
    \paragraph{Joint prediction.} Expansion and localisation are two operations, but we model them as a single conditional distribution by shifting the expansion prediction forward by one level: at level $\ell$ we predict the offsets $C^\ell$ that place the current frontier \emph{and} the expansion states $\Gamma^\ell$ that decide the subset of frontier nodes bifurcating next.
    Given $\tilde T^\ell$, the pair $(C^\ell,\Gamma^\ell)$ determines both $T^\ell$ and $\tilde T^{\ell+1}$, so generation factorises over levels as
    \begin{equation}
        p_\theta(T^0,\dots,T^L) \;=\; \prod_{\ell} p_\theta\big(C^{\ell},\,\Gamma^{\ell} \,\big|\, \tilde T^{\ell}\big).
        \label{eq:shifted_factorisation}
    \end{equation}
    Each factor conditions on the entire partial tree generated so far.
    
    \paragraph{Sampling.} We initialise $T^0=(\{r\},\emptyset,P^0(r)=\mathbf{0})$ and assume the root \emph{has to} expand: $\Gamma^0(r)=1$. 
    To properly model neuron morphologies, we allow $k$ children at the root (soma) with $k$ provided as input. That is, $\tilde T^1$ is given and $\mathcal{A}^1 := \{r_1, r_2, ... , r_k\}$.
    For $\ell\ge 1$, we repeatedly sample positions $C^\ell$ and expansion states $\Gamma^{\ell}$ conditioned on $\tilde T^{\ell}$, defining the tree $T^{\ell}$ as well as $\tilde T^{\ell+1}$ with the new frontier $\mathcal A^{\ell+1}$ according to $\Gamma^{\ell}$. Generation stops when $\Gamma^{\ell}=\mathbf{0}$, i.e. the frontier becomes empty.
    
    \paragraph{Equivariant GNN.} Trees and neurons have an inherent orientation: trees typically grow upward, while neurons are usually oriented perpendicular to the cortical surface \citep{cajal1911}. 
    To encode this inductive bias, we introduce an SO(2)-EGNN model (see Appendix \ref{appendix: SO(2)-EGNN}), inspired by the fully E(n)-equivariant model proposed by \citet{satorras2021n}.
    Every learned quantity inside the network is invariant to rotation about that axis.
    We achieve equivariance by interpreting predicted values in a local frame built from the global principal axis and the incoming branch direction at the parent node.

    \paragraph{Flow Matching.} The prediction of the positions and expansion states at each level $(C^\ell,\Gamma^\ell)$ follows a flow matching model. Flow matching learns a continuous transformation from a simple noise distribution to the distribution of training examples. During training, the network learns the direction in which intermediate noisy samples should move along this transformation. At generation time, we start from noise and follow these learned directions for a small number of steps to obtain a sample (see Appendix \ref{appendix: flowmatching} for more details).
    
    \paragraph{Training sequence.} 
    Let $T$ be a tree with root $r$ and $T_\ell$ the subtree induced by nodes at most \(\ell\) hops from \(r\).
    Training uses the sequence
    \(T_0,\ldots,T_D\), where \(D\) is the maximum depth and $T_0$ only contains $r$.  
    Coarsening \(T_{\ell+1} \to T_{\ell}\) removes one layer of nodes which provide ground-truth expansion targets for $T_\ell$.
    
    \paragraph{Global conditioning.} 
    For generation we consider unconditional as well as class- and tree-conditioned generation.
    For class-conditioning we use a one-hot encoding of the class to condition each flow step and for tree-specific conditioning we rely on a compact topological descriptor of the complete tree.
    The descriptor works by assigning a value to each node
    -- path length, or straight-line distance from the root. Revealing parts of the tree according to the value gives a nested sequence of subgraphs, called a filtration. Zero-dimensional persistent homology records when branches appear and merge in the filtration as points in a persistence diagram. The resulting diagrams are converted to persistence images \citep{adams2017persistence}, concatenated, embedded, and supplied to the network at every flow step (see Appendix \ref{appendix: conditioning}).

    \section{Experiments}
    \label{sec:experiments}

    Our experiments cover both conditional and unconditional generation for both neuron morphologies and botanical trees. We also provide ablations to model size and use the botanical trees to show scaling behaviour with increased tree size.

    \subsection{Setup}
    
    \textbf{Data.}
    For neurons, we use 26{,}469 cortical pyramidal cells from the MICrONS project, collected with \citet{swc_editor} and cleaned following \citet{weis2025unsupervised}. 
    The dataset contains 7 cell-type labels and is split into 22{,}773 training, 2{,}529 validation, and 1{,}167 test cells. In the training split, neurons have 61 nodes on average, a mean depth of 8.2, and a mean soma degree of 7.7. The test split contains the neurons from the manually proofread MICrONS Minnie Column \citet{schneider2025inhibitory} and thus has the highest quality reconstructions.
    \emph{Botanical trees}: We use 3{,}386 QSM reconstructions of broadleaf trees from BioDiv-3DTrees \citep{griese2025large}. The dataset contains six genus labels, and the trees are truncated to depth 10/15/20 and contain 84/195/378 nodes on average. In both datasets, we retain only the root, branch points, and leaves and binarise non-root multifurcations. Further preprocessing details can be found in \cref{app:preprocessing}.
    
    \textbf{Baselines.}
    We compare with SemlaFlow \citep{irwin2024semlaflow} and MorphoGen \citep{zhu2025morphogen}. SemlaFlow is an E(3)-equivariant flow-matching model developed for molecular generation, achieving state-of-the-art performance for molecular generation on QM9 \citep{ramakrishnan2014quantum}. It generates all nodes jointly and predicts their connectivity with a bond head, rather than growing the tree. Unlike our model, it has no distinguished principal axis. 
    MorphoGen \citep{zhu2025morphogen} is the closest neuron-specific generator. It uses a DiT-3D diffusion model that generates the neuron as an unordered point cloud of $2{,}048$ points, from which a tree is recovered post hoc by medial contraction and a greedy growth-guided linking heuristic. It is unconditional only and is handed neither a node count nor the soma degree. 

    We post-process baseline generations to ensure valid trees, effectively selecting the largest connected component (\cref{app:postproc}).
    Our model creates valid trees by construction. We provide SemlaFlow with the reference node count and, for neuronal data, our model with the soma degree.
    
    \textbf{Metrics.} 
    We assess how well generated populations reproduce both the \emph{individual morphological properties} and the \emph{joint distribution} of the reference data. For individual properties, we use 1-Wasserstein ($W_1$) to measure distances between distributions, normalised by the corresponding reference standard deviation ($W_1/\sigma_{\text{ref}}$, see \cref{app:metrics}).
    For joint comparisons, we use (1) a nine-dimensional morphometric vector of graph statistics (\cref{app:metrics}), and (2) TMD persistence images under the path- and radial-from-root filtrations (\cref{appendix: conditioning}). For each representation, we compute MMD$^2$, which measures disagreement between joint distributions of generated and reference samples. To account for finite-sample variation, we subtract the MMD$^2$ between the same reference set and an equally sized, disjoint sample of training morphologies; we report this baseline-adjusted discrepancy as $\Delta \text{MMD}^2$ (closer to zero is better).
    For the morphometric vector, we additionally compute \emph{density} and \emph{coverage}: density indicates whether generated samples lie in regions supported by the reference data, whereas coverage measures how much of the reference distribution they represent. 
    
    \textbf{Reporting conventions.} Samples are drawn with a red circle indicating the root, and branches coloured by depth relative to the projection plane. Structural validity is measured on raw outputs,  morphology metrics use the method-specific post-processing that ensures valid trees.
    
    \subsection{Unconditional Generation}

    \begin{figure}[t]
        \centering
        \includegraphics[width=\textwidth]{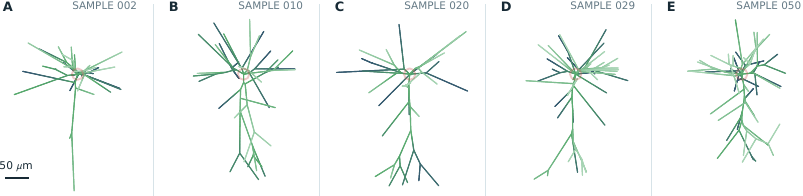}
        \vspace{0.35em}
        \includegraphics[width=\textwidth]{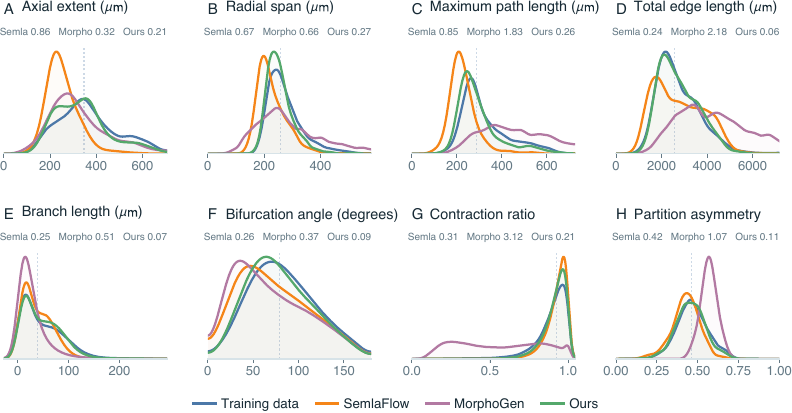}
        \caption{\textbf{Unconditional neuronal morphology generation.} \textbf{Top:} Five representative generations from our model. \textbf{Bottom:} Distributions of eight morphology statistics for the training reference, SemlaFlow, MorphoGen, and our model. Dashed lines mark training-reference medians; grey values give the normalised $W_1$ to the training reference (lower is better). Within-tree distributions weight each morphology equally, and horizontal ranges are determined from the training reference.}
        \label{fig:unconditional-generation}
    \end{figure}

    For unconditional generation in \cref{tab:unconditional-main}, we ask whether the generations are valid morphologies (i.e. graph-theoretic trees) and how closely the generated population matches the reference.
    By design, all our samples are valid trees, as are MorphoGen's given by a spanning tree over the predicted points.
    In contrast, $23.4\%$ of SemlaFlow's raw unconditional outputs are not trees, and $34.0\%$ of the outputs are modified by the postprocessing which contracts degree-2 nodes.
    
    On population statistics, our method outperforms both baselines on the structural properties and on the joint distances to the reference. The density and coverage scores, 0.88 and 0.80 respectively, show that the generated samples lie in regions supported by the reference data and cover much of the reference set.
    MorphoGen performs poorly on both the marginal and the joint evaluations: it over-branches, producing more than twice the reference node count with branches less than half as long. Its density and coverage are near zero, and it has the largest MMD among the evaluated methods. SemlaFlow is closer to the reference than MorphoGen but remains below our model on density, coverage, and every axis-independent marginal statistic (\cref{fig:unconditional-generation}, bottom).

    Our main systematic gap is the under-generation of tall neurons: only 24.2\% of generated cells have an axial extent beyond \(400\,\mu\mathrm{m}\), compared with 33.5\% of the reference cells.
    This recurs in both conditional settings below, although the gap is smaller. 
    A set of samples from our model is shown in \Cref{fig:unconditional-generation}. More extensive samples from Reference, SemlaFlow, MorphoGen, and our model are shown in \cref{fig:appendix_unconditional_gallery} in the appendix.

    \paragraph{Smaller model size.} 
    We experiment with narrower models based on a node-feature width $F$ of 64, 128, and the default 256 in \cref{tab:unconditional-main}.
    We compare $1{,}167$ samples from each method with an equally sized sample of training morphologies.
    Even at $F=64$ with only $1.57$M parameters -- $14\times$ fewer than SemlaFlow and $21\times$ fewer than MorphoGen -- our model still outperforms both.
    Increasing the width improves the joint distances and marginals, while coverage is already saturated at $F=64$.
    
    \begin{table*}[b]
        \centering
        \caption{\textbf{Unconditional neuronal morphology generation.} Distributional metrics on $1{,}167$ samples. The real--real row compares two disjoint training samples of the same size.
        Mean marginal $W_1$ averages the six rotation-invariant statistics defined in \cref{app:metrics}. Default for our method is $F=256$.}
        \label{tab:unconditional-main}
        \resizebox{\textwidth}{!}{%
    \begin{tabular}{lccccccc}
        \toprule
        & & \multicolumn{1}{c}{Structure} & \multicolumn{5}{c}{Distributional agreement} \\
        \cmidrule(lr){3-3}\cmidrule(lr){4-8}
        Method & Params. & \makecell{Valid tree\\(\%) $\uparrow$} & \makecell{Morph.\\$\Delta\mathrm{MMD}^2$ $\downarrow$} & \makecell{TMD\\$\Delta\mathrm{MMD}^2$ $\downarrow$} & \makecell{Coverage\\$\uparrow$} & \makecell{Density\\$\to$ ref.} & \makecell{Mean norm.\\marginal $W_1$ $\downarrow$} \\
        \midrule
        Real--real & \textemdash & 100.0 & 0.0000 & 0.0000 & 0.971 & 0.976 & 0.039 \\
        \midrule
        SemlaFlow & 22.3M & 76.6 & 0.2521 & 0.1038 & 0.223 & 0.155 & 0.389 \\
        MorphoGen & 32.6M & \textbf{100.0} & 0.8436 & 0.2077 & 0.006 & 0.018 & 1.515 \\
        \midrule
        Ours ($F=64$) & 1.57M & \textbf{100.0} & 0.0557 & 0.0146 & \textbf{0.808} & 0.827 & 0.175 \\
        Ours ($F=128$) & 5.69M & \textbf{100.0} & 0.0471 & 0.0108 & 0.805 & 0.862 & 0.163 \\
        Ours (default) & 21.6M & \textbf{100.0} & \textbf{0.0393} & \textbf{0.0092} & 0.798 & \textbf{0.876} & \textbf{0.133} \\
        \bottomrule
    \end{tabular}%
}

    \end{table*}

    \subsection{Class-Conditioned Generation}

    \begin{figure}[!htbp]
        \centering
        \includegraphics[width=0.95\textwidth]{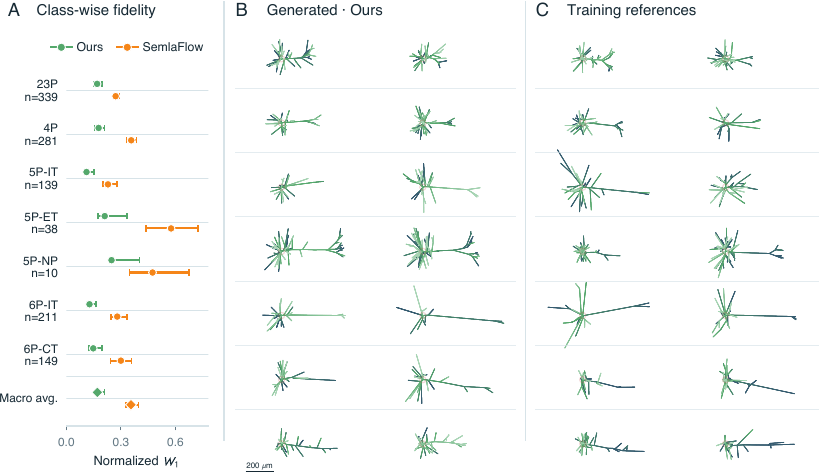}
        \caption{\textbf{Class-conditioned neuronal morphology generation.} \textbf{(A)} Class-wise morphology discrepancy for our model and SemlaFlow relative to the training trees. The values represent normalised 1-Wasserstein distance, averaged over five rotation-invariant statistics (lower is better). Whiskers show 95\% bootstrap intervals, resampling the generated morphologies while holding the sampled training reference fixed. $n$ is the number of generated morphologies. \textbf{(B)} Two selected generations from our model per class. \textbf{(C)} Two selected training-set references per class.}
        \label{fig:class_conditioned_samples}
    \end{figure}

    Conditioning on the seven cell-type labels yields the expected class differences (\cref{fig:class_conditioned_samples}): compact, bushy arbours for 23P and 4P cells, tall apical trunks for 5P-ET and 5P-NP, and long, sparsely branched layer-6 cells shown next to reference cells.

    The class structure also holds up quantitatively. 
    We evaluate maximum path length, branch length, bifurcation angle, contraction, and partition asymmetry. 
    Node count is excluded because it is supplied to SemlaFlow, while directional extents are excluded because SemlaFlow has no fixed vertical axis.
    Averaged over the seven classes and five statistics, the normalised $W_1$ is $0.170$ for our model and $0.355$ for SemlaFlow. 
    MorphoGen only supports unconditional generation and is thus not included.
    \cref{fig:appendix_class_morphology} reports all statistics for each class separately. For our method, within-tree statistics such as branch length and bifurcation angle are matched within one standard deviation in every class.

    \subsection{Morphology-Guided Generation}

    \begin{figure}[ht]
        \centering
        \includegraphics[width=\textwidth]{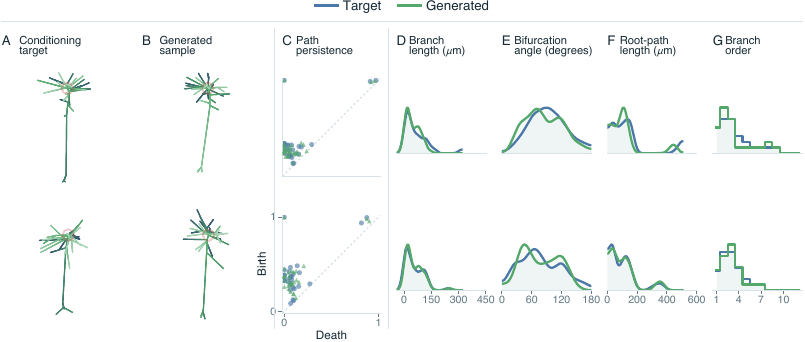}
        \vspace{0.35em}
        \includegraphics[width=\textwidth]{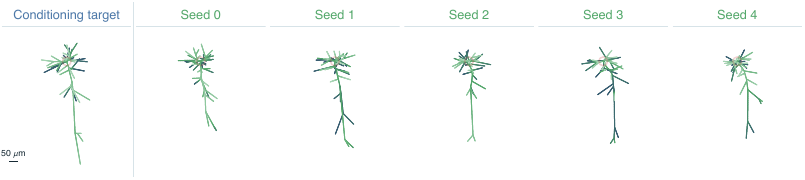}
        \caption{\textbf{TMD-conditioned generation.} \textbf{Top:} Each row shows a conditioning target, one generated sample, their path-based persistence diagrams, and within-tree distributions of branch length, bifurcation angle and branch order. \textbf{Bottom:} 5 independent samples conditioned on the same target.}
        \label{fig:conditional_samples}
    \end{figure}

    \begin{table*}[b]
        \centering
        \caption{\textbf{TMD-conditioned generation.} Each method generates one morphology for each of 1{,}167 held-out targets. Entries in the \textit{persistence diagrams} and \textit{within-tree distributions} are means over target--generation pairs. The former compares normalised persistence diagrams, and the latter compares distributions of within-tree attributes including branch length, sibling angle, and branch order. The reference row pairs each target with a different held-out reference using a fixed-seed random derangement, providing a no-conditioning baseline.}
        \label{tab:tmd-conditional-main}
        \resizebox{\textwidth}{!}{%
    \begin{tabular}{lcccccc}
        \toprule
        & \multicolumn{1}{c}{Validity} & \multicolumn{2}{c}{Persistence diagrams} & \multicolumn{3}{c}{Within-tree distributions} \\
        \cmidrule(lr){2-2}\cmidrule(lr){3-4}\cmidrule(lr){5-7}
        Method & \makecell{Valid tree\\(\%) $\uparrow$} & \makecell{Path PD\\$W_1$ $\downarrow$} & \makecell{Radial-root PD\\$W_1$ $\downarrow$} & \makecell{Branch length\\$W_1$ ($\mu$m) $\downarrow$} & \makecell{Sibling angle\\$W_1$ (${}^\circ$) $\downarrow$} & \makecell{Branch order\\$W_1$ $\downarrow$} \\
        \midrule
        Reference & 100.0 & 5.914 & 6.453 & 16.15 & 8.34 & 1.190 \\
        SemlaFlow & 63.2 & 3.158 & \textbf{2.453} & 18.01 & 10.61 & 0.668 \\
        Ours & \textbf{100.0} & \textbf{2.939} & 2.803 & \textbf{13.38} & \textbf{9.50} & \textbf{0.548} \\
        \bottomrule
    \end{tabular}%
}

    \end{table*}

    Our method allows us to condition directly on the morphology of a given tree based on the fixed-length TMD vector.
    \Cref{fig:conditional_samples} shows two TMD-conditioned generations from our model compared to the ground-truth morphology. The generations retain the target's overall shape, with a long main branch and most branching near the top. Their path persistence diagrams, which the TMD condition directly represents (see \Cref{appendix: conditioning}), align closely. The distributions of branch length, distance from the root, and branch order also agree well, while branching angles have slightly larger differences. This suggests that TMD primarily controls overall structure rather than finer geometric details. The bottom of \cref{fig:conditional_samples} illustrates this with five generations from the same target, retaining the overall structure while differing in smaller details. See \cref{fig:appendix_tmd_multiseed_atlas} for more examples. 
    
    \Cref{tab:tmd-conditional-main} evaluates TMD-conditioned generation across the full test set. Our method produces a valid tree in every case, compared to $63.2\%$ for SemlaFlow, and gives lower errors for path persistence, branch length, branching angle. SemlaFlow performs slightly better on radial persistence. Both persistence-diagram distances are roughly halved relative to a randomly paired held-out tree (top row). This indicates that the generations follow the specified target rather than simply matching the overall population. Appendix \Cref{tab:tmd-multiseed} provides a stricter comparison, where conditional generations are compared to the target and a root-degree matched random sample. Averaged over five seeds, the target is closer in about $91\%$ of comparisons. In $97.9\%$ of cases, our generator produces five unique generations out of the five seeds.
    
    \subsection{Botanical Trees}

    We evaluate unconditional and TMD-conditioned generation on botanical trees (\cref{tab:botanical-generation-comparison}). Our model gives the lower discrepancy on all four morphometrics in both regimes, with a mean normalised $W_1$ of $0.236$ unconditionally and $0.141$ under TMD guidance -- between $2.8\times$ and $4.0\times$ better than SemlaFlow. SemlaFlow degrades on validity: at most $12.5\%$ of its raw outputs are trees at all, against $76.6\%$ on neurons. As with neurons, morphology guidance is the strongest regime. Performance is therefore maintained across two structurally and biologically distinct domains, supporting frontier expansion as a general model of branching structure rather than one tuned to a single morphology.

    \paragraph{Scaling with tree size.}

    \begin{figure}[t]
        \centering
        \includegraphics[width=\textwidth]{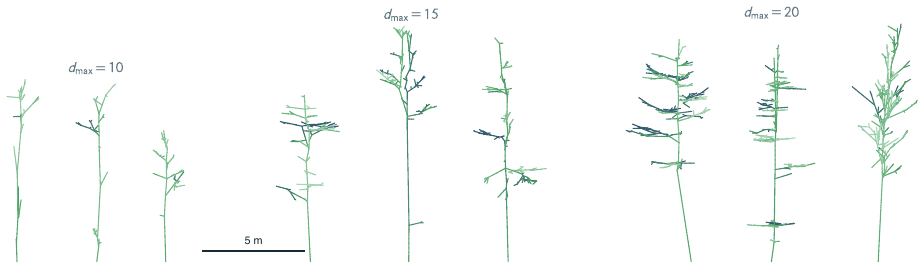}
        \caption{\textbf{Generated botanical trees across depth caps.} Three representative samples are shown for each maximum topological depth $d_{\max}\in\{10,15,20\}$.}
        \label{fig:botanical-depth-gallery}
    \end{figure}
    \begin{table*}[!htpb]
    \centering
    \caption{\textbf{Botanical-tree generation.} Population-level discrepancies on botanical trees restricted to depth 10 (full dataset). Morphology entries are tree-balanced 1-Wasserstein distances normalised by the reference standard deviation $W_1/\sigma_{\mathrm{ref}}$ (lower is better). Mean averages the distances. Bold marks the better result per training regime.}
    \label{tab:botanical-generation-comparison}
    \setlength{\tabcolsep}{3pt}
    \begin{tabular*}{\textwidth}{@{\extracolsep{\fill}}llrrrrrr@{}}
        \toprule
        Regime & Method & \makecell{Valid tree\\(\%) $\uparrow$} & \makecell{Max. path\\length $\downarrow$} & \makecell{Branch\\length $\downarrow$} & \makecell{Bifurcation\\angle $\downarrow$} & \makecell{Partition\\asymmetry $\downarrow$} & \makecell{Mean\\$\downarrow$} \\
        \midrule
        \multirow{2}{*}{Unconditional} & SemlaFlow & 3.9 & 1.280 & 0.245 & 0.297 & 0.822 & 0.661 \\
         & Ours & \textbf{100.0} & \textbf{0.443} & \textbf{0.167} & \textbf{0.177} & \textbf{0.157} & \textbf{0.236} \\
        \addlinespace[2pt]
        \multirow{2}{*}{TMD-conditioned} & SemlaFlow & 12.5 & 0.982 & 0.172 & 0.417 & 0.683 & 0.564 \\
         & Ours & \textbf{100.0} & \textbf{0.165} & \textbf{0.087} & \textbf{0.238} & \textbf{0.076} & \textbf{0.141} \\
        \bottomrule
    \end{tabular*}
\end{table*}

    \Cref{fig:botanical-depth-gallery} illustrates the increasingly detailed structures generated at maximum depths of 10, 15, and 20 hops from the root. For the quantitative comparison, we evaluate unconditional generation on capped test sets (\cref{tab:botanical-depth-scaling}). To make the largest set feasible for SemlaFlow, we exclude the same 199 trees exceeding 1{,}110 nodes at every depth. Our model gives the lower discrepancy on all four morphometrics at every depth: its mean normalised $W_1$ ranges from $0.27$ at D10 to $0.38$ at D20, whereas for SemlaFlow it ranges from $0.66$ to $1.84$. Sampling time grows from $0.192$ to $0.617$ seconds per tree for our method, compared with $0.562$ to $12.317$ seconds for SemlaFlow.

    \begin{table*}[!htbp]
    \centering
    \caption{\textbf{Unconditional botanical-tree generation across depth caps.} Columns are $W_1/\sigma_{\mathrm{ref}}$ discrepancies (lower is better), Mean averages the four discrepancies, and Time is seconds per tree. Bold marks the better result within each depth.}
    \label{tab:botanical-depth-scaling}
    \setlength{\tabcolsep}{4pt}
    \begin{tabular*}{\textwidth}{@{\extracolsep{\fill}}llrrrrrr@{}}
        \toprule
        Depth & Method & \makecell{Max. path\\length $\downarrow$} & \makecell{Branch\\length $\downarrow$} & \makecell{Bifurcation\\angle $\downarrow$} & \makecell{Partition\\asymmetry $\downarrow$} & \makecell{Mean\\ $\downarrow$} & \makecell{Time\\(s/tree)} \\
        \midrule
        \multirow{2}{*}{D10} & SemlaFlow & 1.314 & 0.275 & 0.330 & 0.737 & 0.664 & 0.562 \\
         & Ours & \textbf{0.542} & \textbf{0.178} & \textbf{0.197} & \textbf{0.148} & \textbf{0.266} & \textbf{0.192} \\
        \addlinespace[2pt]
        \multirow{2}{*}{D15} & SemlaFlow & 2.347 & 0.441 & 0.272 & 1.871 & 1.233 & 3.714 \\
         & Ours & \textbf{0.639} & \textbf{0.227} & \textbf{0.200} & \textbf{0.139} & \textbf{0.301} & \textbf{0.390} \\
        \addlinespace[2pt]
        \multirow{2}{*}{D20} & SemlaFlow & 2.572 & 0.361 & 0.452 & 3.964 & 1.837 & 12.317 \\
         & Ours & \textbf{0.702} & \textbf{0.250} & \textbf{0.198} & \textbf{0.366} & \textbf{0.379} & \textbf{0.617} \\
        \bottomrule
    \end{tabular*}
\end{table*}

    \section{Conclusion}
    
    Autoregressive frontier expansion generates 3D tree structures by predicting topology and geometry one level at a time. On cortical neurons and botanical trees, it produces valid trees by construction, matches reference morphology distributions more closely than the evaluated baselines, and scales to larger trees. Conditioning on cell type or a topological description enables class-level and target-specific control. These results indicate that iterative tree construction is a useful inductive bias for generative models of branching structures. Further questions include generation in the context of neighbouring trees and neurons, which the existing conditioning targets do not yet cover.

    \subsection*{AI use statement}
    We used large language models for language editing, including spell-checking, condensing, and clarifying text. We also used AI tools for coding assistance, including implementing, debugging, and testing research and evaluation code. All AI-assisted work was reviewed by the authors. The research ideas, experimental design, and interpretation of the results are entirely our own.
    
    \section{Acknowledgements}
    The authors acknowledge the financial support by the Federal Ministry of Research, Technology and Space of Germany and by Sächsische Staatsministerium für Wissenschaft, Kultur und Tourismus in the programme Center of Excellence for AI-research ``Center for Scalable Data Analytics and Artificial Intelligence Dresden/Leipzig'', project identification number: ScaDS.AI
    
    \bibliographystyle{iclr2027_conference}
    \bibliography{ref}
    
    \newpage
\appendix

\section*{Appendix}

The appendix includes pseudocode for our method (\cref{app:pseudocode}), details on data preprocessing (\cref{app:preprocessing}), details on the flow-matching procedure (\cref{appendix: flowmatching}), the SO(2)-equivariant architecture (\cref{appendix: SO(2)-EGNN}), our conditioning vector (\cref{appendix: conditioning}), metrics (\cref{app:metrics}), postprocessing for our baselines (\cref{app:postproc}), as well as additional results and samples (\cref{app:additional-quantitative,app:additional-samples}).

\section{Pseudocode for Autoregressive Frontier Expansion}
\label{app:pseudocode}

\begin{algorithm}[H]
    \DontPrintSemicolon
    \SetAlgoLined
    \SetAlgoNlRelativeSize{0}
    \SetKwIF{If}{ElseIf}{Else}{if}{:}{else if}{else}{}
    \SetKwFor{For}{for}{:}{}
    \SetKwFor{While}{while}{:}{}
    \SetKwFor{ForEach}{foreach}{:}{}
    \KwIn{root degree $k$}
    $V^1 = \{r, r_1, \dots, r_k\};\; E^1 = \{(r, r_i)\}_{i=1}^{k};\; P^0(r) = \mathbf 0$ \tcp*[r]{root at the origin}
    $\mathcal{A}^1 = \{r_1, \dots, r_k\}$; $\ell = 0$ \tcp*[r]{$k$ unplaced children: this is $\tilde T^1$}

    \While{$\mathcal{A}^{\ell+1} \neq \emptyset$}{ 
        $\ell = \ell + 1$\;
        Sample $(C^\ell, \Gamma^\ell) \sim p_\theta(\,\cdot \mid \tilde T^\ell)$ \tcp*[r]{flow matching, Appendix~\ref{appendix: flowmatching}}

        $P^{\ell} = P^{\ell-1}$\;
        \ForEach{$v \in \mathcal{A}^\ell$}{
            $P^{\ell}(v) = P^{\ell-1}(\pi(v)) + C^{\ell}(v)$ \tcp*[r]{localise: $\tilde T^\ell \to T^\ell$}
        }

        $V^{\ell+1} = V^\ell;\; E^{\ell+1} = E^\ell;\; \mathcal{A}^{\ell+1}=\emptyset$\;
        \ForEach{$v \in \mathcal{A}^\ell$ with $\Gamma^\ell(v) = 1$}{
            $V^{\ell+1} = V^{\ell+1} \cup \{v_L, v_R\};\; E^{\ell+1} = E^{\ell+1} \cup \{(v,v_L), (v,v_R)\}$\;
            $\mathcal{A}^{\ell+1} = \mathcal{A}^{\ell+1} \cup \{v_L, v_R\}$ \tcp*[r]{expand: $T^\ell \to \tilde T^{\ell+1}$}
        }
    }
    \Return{$T^\ell$}
    \caption{Tree generation procedure}
    \label{algo:generation}
\end{algorithm}

Pseudocode for our autoregressive tree generation. 
The root is initialised with $k$ unplaced children, where $k$ is an input to the sampler. In each iteration the offsets $C^\ell$ and expansion states $\Gamma^\ell$ of the current frontier $\mathcal{A}^\ell$ are sampled jointly given the intermediary tree $\tilde T^\ell$.
The offsets place the frontier relative to its parents, and every node with $\Gamma^\ell(v)=1$ receives two children, which form the next frontier. The process continues until no node expands and the frontier is empty.

\section{Preprocessing and Tree Representation} \label{app:preprocessing}

We assume trees to consist of binary branching nodes and leaves only. 
We thus preprocess all trees to their branching skeletons by contracting all degree-2-paths such that only branching nodes and leaves remain.
Degree-2-paths encode the local shape of branches rather than structure.
We leave the extension of our method to also predict those local branch shapes as future work.
Performing this simplification also benefits GNNs and message passing: contracting degree-2 paths shortens the graph distances that message passing must span.  In an unreduced MICrONS tree, even a 10-layer GNN may fail to propagate information from one branching node to the next.

We further assume that apart from the root (to be able to generate neuron morphologies, see next paragraph) trees are binary, which simplifies the decision space for the learned expansion target. A three-way junction is split losslessly by inserting a node at the intersection; at wider junctions we keep the two thickest children and drop the rest, which removes roughly $1.25\%$ of neuron nodes.
Handling junctions with more than two children natively is kept for future work.

The root is exempt from this constraint: a soma carries $1$ to $K_{\max}$ primary dendrites, and we set $K_{\max}=23$, the largest number observed in the corpus, so that no neuron is excluded by the bound. In our data the mean soma degree is $7.7$. 
Botanical trees have root degree $1$ or $2$, the base of the trunk being a single cylinder. The rank of each root child among its siblings is supplied to the model as an input feature; \cref{app:so2-root} explains why it is needed -- all root children share a single local frame, so rank is the only signal that distinguishes them.

For the scaling experiment, we use the botanical trees as they are much larger and deeper than neurons, reaching 70 hops and 3k nodes on average (max 326 hops and 27k nodes), whereas neurons have 8 hops and 61 nodes (max 35 hops and 537 nodes).
We thus use the first 10/15/20 hops of each of the trees and only consider trees where size at 20 hops does not exceed 1{,}110 nodes to be able to run SemlaFlow. The cap removes around 6\% of trees which would otherwise again have up to 3k nodes, and now reaches 378 nodes on average. In contrast to the full D10 trees used for unconditional generation in \cref{tab:botanical-generation-comparison} with an average number of nodes of 84, the cap globally dropped the densest trees such that the average number of nodes in \cref{tab:botanical-depth-scaling} for D10 is 78.

\section{Flow Matching} \label{appendix: flowmatching}

At each expansion step, the newly created frontier $\mathcal{A}^{\ell}$ consists of leaves whose coordinates and \emph{next-step} expansion state must be generated. 
As described in the main paper, both predictions are performed jointly using flow matching.
Concretely, we parametrise this as a conditional flow matching problem \citep{lipman2022flow, liu2022flow} on the \emph{frontier state}, 
\( X^{\ell}=[\check C^{\ell},\,\Gamma^{\ell}] \in \mathbb{R}^{|\mathcal{A}^{\ell}|\times 4},\) 
conditioned on the full intermediate expanded tree $\tilde T^{\ell}$. For flow matching we treat $\Gamma$ as continuous.

In this section, we drop the tree expansion level $\ell$ from the superscript to make the notation easier to read.
In the subscript we now write the \emph{flow time} $t \in [0,1]$, so $P^{\ell}_t(v)$ is the position of node $v$ in round $\ell$ at flow time $t$.
Note that $t$ runs from noise to data: $t=0$ is a draw from the prior and $t=1$ is the data.
The offsets $C(v)$ of \cref{eq:relative_position_parametrisation} are expressed in world axes for improved readability. In practice, our method works in \emph{local coordinates} $\check C(v)$ to facilitate learning and to incorporate SO(2)-equivariance, see \cref{app:so2-frames,app:so2-decoding}.

\paragraph{Probability path.}
We use the linear (optimal-transport) interpolant between a prior sample $X_0$ and the data $X_1$:
\begin{equation}
    X_t = (1-t)\,X_0 + t\,X_1, \qquad X_0\sim\mathcal{N}(0,\Sigma),\qquad t\sim\mathcal{U}[0,1],
    \label{eq:ot_path}
\end{equation}
drawn independently for each frontier node $v$. 
Here $\Sigma$ is diagonal and shared by every node: it has unit variance on the expansion coordinate, and per-axis scales set to the spread of the ground-truth offsets along each frame axis, in each corpus's normalised coordinates.
In training we sample a single $t$ per graph.
Thus, each training example consists of a tree $T$ sampled at some level $\ell$ together with a flow time $t$, and we aim to predict the velocity of the path at $X_t$.

\paragraph{Velocity field.}
Because the path in \cref{eq:ot_path} is a straight line, its velocity is constant along the path and given in closed form by
\begin{equation}
    u(X_t,t) \;=\; \frac{\mathrm{d}X_t}{\mathrm{d}t} \;=\; X_1 - X_0 .
    \label{eq:velocity_target}
\end{equation}

\paragraph{Network and Inputs.}
Our velocity network is the SO(2)-EGNN model. It reads node coordinates only to form the rotation- and translation-invariant edge quantities of \cref{app:so2-edges}, and never updates them. We construct $P_t$ by overwriting only the frontier node coordinates:
\begin{equation}
    P_t(v) =
    \begin{cases}
    P_1(v), & v\notin \mathcal{A},\\
    P_1(\pi(v)) + C_t(v), & v\in \mathcal{A}.
    \end{cases}
    \label{eq:pt_construction}
\end{equation}
All non-frontier nodes are already placed and provide geometric context. $\check C_t(v)$ is converted back to the global frame as input to the invariant network. However, note that $P_t$ is a transient input, not part of the integrated state: it is rebuilt from $P_1$ at every step and discarded, while the state itself never leaves frame coordinates.

Let $H$ denote deterministic graph inputs: node features and constructed geometric edge features. We provide $t$ as a broadcast node feature and use the interpolated expansion state $\Gamma_t$ as an additional node feature on frontier nodes (and $0$ elsewhere). Then
\begin{equation}
    v_\theta:\ (P_t, H, \Gamma_t, t) \mapsto \widehat{V},
\end{equation}
giving velocity predictions on the frontier state, $\widehat{V}=[\widehat{V}_{\check C},\widehat{V}_\Gamma]$. Here $\widehat{V}_{\check C}(v)$ is a triple of frame coordinates, in the same basis as $\check C_t(v)$, and $\widehat{V}_\Gamma(v)$ is a scalar.

\paragraph{Training Objective.}
We regress the velocity with an MSE loss on the frontier nodes only.
Let $\theta$ denote the parameters of the SO(2)-EGNN.
\begin{equation}
    \mathcal{L}(\theta)=
    \mathbb{E}_{t,X_0}\!\left[
    \sum_{v\in\mathcal{A}}\big\|\widehat{V}_{\check C}(v)-\big(\check C_{1}(v)-\check C_{0}(v)\big)\big\|_2^2
    +\sum_{v\in\mathcal{A}}\big(\widehat{V}_\Gamma(v)-\big(\Gamma_{1}(v)-\Gamma_{0}(v)\big)\big)^2
    \right],
\end{equation}
where the ground-truth target $\check C_1(v)$ is the observed offset read off in the local frame at $v$. Prediction and target are thus in a common basis and the loss needs no change of coordinates.
The binary expansion label is mapped to $\Gamma_1 \in \{-1,+1\}$ rather than $\{0,1\}$, so that it is symmetric about the prior mean; the decision threshold at sampling time is therefore $0$.

\paragraph{Sampling.}
At inference we draw $X_0$ from the prior and integrate \cref{eq:velocity_target} forward with explicit Euler steps on a uniform grid $t_k = k/K$,
\begin{equation}
    X_{t_{k+1}} = X_{t_k} + \tfrac{1}{K}\,v_\theta(P_{t_k},H,\Gamma_{t_k},t_k),
    \qquad k=0,\dots,K-1,
\end{equation}
with $K=10$ steps.
\Cref{algo:flowsampling} gives the procedure in full.
We note that for the algorithm, the integrated state $X$ stays in frame coordinates throughout; the frame is used only to hand the network world coordinates to measure, and to place the node at the end.
It is built once from the already-placed parent positions and held fixed across all $K$ steps, so the basis its inputs are measured in and the basis its outputs are decoded in are the same, and neither drifts with the state being integrated (\cref{app:so2-frames}). Only frontier-adjacent geometry is recomputed between steps.

\begin{algorithm}[t]
    \DontPrintSemicolon
    \SetAlgoLined
    \SetAlgoNlRelativeSize{0}
    \SetKwIF{If}{ElseIf}{Else}{if}{:}{else if}{else}{}
    \SetKwFor{For}{for}{:}{}
    \SetKwFor{ForEach}{foreach}{:}{}
    \KwIn{intermediate tree $\tilde T^{\ell}$, frontier $\mathcal A$, steps $K$}
    local frames $(\mathbf f_v,\mathbf s_v,\hat{\mathbf u}) \leftarrow$ from placed parents \tcp*[r]{built once, \cref{app:so2-frames}}
    $X \sim \mathcal{N}(0, \Sigma)$ \tcp*[r]{prior sample in frame coords, \cref{eq:ot_path}}
    \For{$k = 0$ \KwTo $K-1$}{
        $t_k = k/K$\;
        \ForEach{$v \in \mathcal{A}$}{
            $P_{t_k}(v) = P(\pi(v)) + C_{t_k}(v)$ \tcp*[r]{decode $X_{\check C}$, \cref{eq:frame_coords}}
        }
        $H \leftarrow$ recompute branch angles, edge features for $\mathcal A$ \tcp*[r]{frames locked}
        $\widehat V = v_\theta(P_{t_k}, H, X_\Gamma, t_k)$\;
        $X = X + \tfrac{1}{K}\widehat{V}$ \tcp*[r]{explicit Euler, in frame coords}
    }
    \ForEach{$v \in \mathcal{A}$}{
        $P^{\ell}(v) = P(\pi(v)) + C(v)$ \tcp*[r]{final placement}
    }
    \Return{$P^{\ell}$, \quad $\Gamma^{\ell} = \mathbf{1}[X_\Gamma > 0]$} \tcp*[r]{to build $\tilde T^{\ell+1}$}
    \caption{Flow-matching sampling of the frontier state}
    \label{algo:flowsampling}
\end{algorithm}

\section{SO(2)-EGNN} \label{appendix: SO(2)-EGNN}

\subsection{Invariant edge features} \label{app:so2-edges}

The original EGNN achieves E(n) equivariance by using only pairwise squared distances as geometric input to edge messages, deliberately avoiding directional information that would break rotational symmetry \citep{satorras2021n}. For structures with a preferred axis, we instead decompose displacements into axis-parallel and perpendicular components.

For each directed edge \((i\!\to\! j)\) with coordinates \(\mathbf{x}_i,\mathbf{x}_j\), we compute the relative displacement \(\mathbf{r}_{ij}=\mathbf{x}_j-\mathbf{x}_i\) and decompose it with respect to a rotation axis \(\hat{\mathbf u}\), fixed per corpus.
We form the axial component \(d^{\parallel}_{ij}=\mathbf{r}_{ij}^{\top}\hat{\mathbf u}\) and the perpendicular residual \(\mathbf{r}^{\perp}_{ij}=\mathbf{r}_{ij}-d^{\parallel}_{ij}\hat{\mathbf u}\), then take \(\rho_{ij}=\|\mathbf{r}^{\perp}_{ij}\|_2\). 
The pair \((\rho_{ij}, d^{\parallel}_{ij})\) is invariant to rotations around \(\hat{\mathbf u}\). 
We enrich these scalars with two branch angles, shared by both directions of the tree edge between $v$ and its parent. Let $\mathbf{r}_v = P(v) - P(\pi(v))$ be the branch, $\hat{\mathbf r}^{\perp}_v$ its unit projection onto the plane perpendicular to $\hat{\mathbf u}$, and $(\mathbf{f}_v, \mathbf{s}_v, \hat{\mathbf u})$ the local frame of \cref{app:so2-frames}. Then
\begin{equation}
    \cos\psi_v = \mathbf{f}_v^{\top}\hat{\mathbf r}^{\perp}_v, \qquad
    \sin\psi_v = \mathbf{s}_v^{\top}\hat{\mathbf r}^{\perp}_v, \qquad
    \cos\phi_v = \frac{\mathbf{r}_v^{\top}\hat{\mathbf u}}{\|\mathbf{r}_v\|_2},
    \label{eq:branch_angles}
\end{equation}
so $\psi_v$ is the azimuth of the branch about $\hat{\mathbf u}$, measured from the incoming direction at the parent, and $\phi_v$ its tilt from the axis. Both are invariant to rotations about $\hat{\mathbf u}$, since the frame co-rotates with the branch.

\subsection{Message passing} \label{app:so2-mp}

Every tree edge is used in both directions. Writing $\mathbf{e}_{ij}$ for the invariant edge features above, one layer updates the node features $\mathbf{h}$ as
\begin{equation}
    \mathbf{m}_{ij} = \mathrm{MLP}_e\big(\mathbf{h}_i, \mathbf{h}_j, \mathbf{e}_{ij}\big), \qquad
    \mathbf{m}_j = \sum_{(i\to j)\in E} \mathbf{m}_{ij}, \qquad
    \mathbf{h}_j \leftarrow \mathbf{h}_j + \mathrm{MLP}_h\big(\mathrm{LN}(\mathbf{h}_j), \mathbf{m}_j\big),
    \label{eq:so2_layer}
\end{equation}
where $\mathrm{MLP}_e, \mathrm{MLP}_h$ are two-layer perceptrons and $\mathrm{LN}$ is layer normalisation. This is the EGNN layer of \citet{satorras2021n} with the squared distance replaced by $\mathbf{e}_{ij}$ and the coordinate update removed: coordinates are read once to form $\mathbf{e}_{ij}$ and never modified. Since messages only travel along tree edges, we precede every second layer with an induced-set attention block \citep{lee2019set}, in which a small set of learned per-graph tokens attends to all node features and the nodes attend back; it acts on invariant features only, so invariance is preserved.

\subsection{Local frames} \label{app:so2-frames}

For a node $v$ whose parent $\pi(v)$ itself has a parent $\pi(\pi(v))$, let $\mathbf{w}_v = P(\pi(v)) - P(\pi(\pi(v)))$ be the incoming branch direction at the parent. Projecting it into the plane perpendicular to $\hat{\mathbf u}$ and normalising gives
\begin{equation}
    \mathbf{f}_v = \frac{\mathbf{w}_v - (\mathbf{w}_v^{\top}\hat{\mathbf u})\,\hat{\mathbf u}}
                        {\|\mathbf{w}_v - (\mathbf{w}_v^{\top}\hat{\mathbf u})\,\hat{\mathbf u}\|_2},
    \qquad
    \mathbf{s}_v = \hat{\mathbf u} \times \mathbf{f}_v ,
    \label{eq:local_frame}
\end{equation}
so that $\mathbf{f}_v \perp \hat{\mathbf u}$ by construction. We call $(\mathbf{f}_v, \mathbf{s}_v, \hat{\mathbf u})$ the \emph{local frame} at $v$: an in-plane forward direction inherited from the parent branch, a sideways direction completing the perpendicular plane, and the corpus axis. A triple $\check{\mathbf c}$ of coordinates in this frame denotes the world vector
\begin{equation}
    C(v) = \check c_1\,\mathbf{f}_v + \check c_2\,\mathbf{s}_v + \check c_3\,\hat{\mathbf u}.
    \label{eq:frame_coords}
\end{equation}
For trees, the basis always contains up (the principal axis). The first horizontal axis aligns with the parent-branch direction (projected to the horizontal plane), and the last axis completes the orthonormal frame.

\subsection{Equivariance} \label{app:so2-decoding}

Because the features of \cref{app:so2-edges} are all invariant, so is everything the network computes from them: the head emits a triple $(a_v, b_v, c_v)$ of \emph{frame} coordinates, together with the expansion velocity $\gamma_v$. Equivariance is supplied by the per-node local frame, against which \cref{eq:frame_coords} decodes those coordinates into geometry.

Let $Q$ be a rotation about $\hat{\mathbf u}$, applied to every node coordinate. The edge quantities of \cref{app:so2-edges} are unchanged by construction, hence so are all node features and the head outputs. The prior of \cref{eq:ot_path} is defined in frame coordinates with $\Sigma$ fixed, so it carries no dependence on orientation either, and the Euler update of \cref{algo:flowsampling} adds invariant increments to an invariant state: the integrated state $X$ is therefore invariant. The frame vectors, by contrast, are built from coordinates and co-rotate, $\mathbf{f}_v \mapsto Q\mathbf{f}_v$ and $\mathbf{s}_v \mapsto Q\mathbf{s}_v$, with $\hat{\mathbf u}$ fixed by $Q$. Placement decodes that invariant state against the co-rotating frame (\cref{eq:frame_coords}), so each offset maps to $Q\,C(v)$; with the root at the origin, the generated positions co-rotate with the input.

\subsection{The root} \label{app:so2-root}

The root has no grandparent, so \cref{eq:local_frame} has no incoming direction to work with, and its $k$ children are created in a single step rather than one at a time ($1 \le k \le K_{\max}$). To resolve both facts, \textbf{all $k$ root children share a single frame}, whose forward direction $\mathbf{f}_0$ is aligned during training with the child whose \emph{entire subtree} reaches furthest along $-\hat{\mathbf u}$. At sampling no child is yet placed, so $\mathbf{f}_0$ is instead a single azimuth drawn uniformly at random per root; its absolute value is precisely the SO(2) degree of freedom the model does not fix. With the frame shared, no geometric signal distinguishes one root child from another -- their frames are identical and their parent is the same node. We therefore supply each root child's rank among its siblings as a $K_{\max}$-wide one-hot node feature.

\section{Topological Morphological Conditioning} \label{appendix: conditioning}

\begin{figure}
    \centering
    \resizebox{0.25\linewidth}{!}{\begin{tikzpicture}
  \draw[tg/edge] (0.0,0.0) -- (0.18117333157176452,0.6761480784023478);
  \draw[tg/edge] (0.18117333157176452,0.6761480784023478) -- (-0.22414386804201342,1.0953353488403286);
  \draw[tg/edge] (0.18117333157176452,0.6761480784023478) -- (0.6072956374218469,1.4937175130420053);
  \draw[tg/edge] (0.0,0.0) -- (1.521457654690411,1.0417133026816707);
  \draw[tg/edge] (1.521457654690411,1.0417133026816707) -- (1.61297343904537,2.15599552062015);
  \draw[tg/edge] (1.521457654690411,1.0417133026816707) -- (2.1338300903862057,1.3952666932749445);
  \node[tg/node] (A-0) at (0.0,0.0) {$ r $};
  \node[tg/node] (A-1) at (0.18117333157176452,0.6761480784023478) {$ v_1 $};
  \node[tg/node] (A-2) at (-0.22414386804201342,1.0953353488403286) {$ v_2 $};
  \node[tg/node] (A-3) at (0.6072956374218469,1.4937175130420053) {$ v_3 $};
  \node[tg/node] (A-4) at (1.521457654690411,1.0417133026816707) {$ v_4 $};
  \node[tg/node] (A-5) at (1.61297343904537,2.15599552062015) {$ v_5 $};
  \node[tg/node] (A-6) at (2.1338300903862057,1.3952666932749445) {$ v_6 $};
\end{tikzpicture}}%
    \hspace{0.7cm}%
    \resizebox{0.33\linewidth}{!}{\begin{tikzpicture}[x=0.8cm, y=0.6cm]
  \draw[->, thick] (0,-0.5) -- (3.462,-0.5);
  \draw[thick] (0,-0.5) -- (0,1.800);
  \node[below, yshift=-10pt, font=\scriptsize] at (1.731,-0.5) {filtration (tree distance)};

  \draw[thick] (0,-0.5) -- (0,-0.62);
  \node[below, font=\scriptsize] at (0,-0.62) {0};
  \draw[thick] (1,-0.5) -- (1,-0.62);
  \node[below, font=\scriptsize] at (1,-0.62) {1};
  \draw[thick] (2,-0.5) -- (2,-0.62);
  \node[below, font=\scriptsize] at (2,-0.62) {2};
  \draw[thick] (3,-0.5) -- (3,-0.62);
  \node[below, font=\scriptsize] at (3,-0.62) {3};

  \node[left, font=\scriptsize] at (0,0.000) {$(r,v_5)$};
  \draw[line width=1.5pt, orange!80!black] (0.000000,0.000) -- (2.961943,0.000);
  \fill[orange!80!black] (0.000000,0.000) circle (1.5pt);
  \fill[orange!80!black] (2.961943,0.000) circle (1.5pt);
  \node[left, font=\scriptsize] at (0,0.600) {$(v_4,v_6)$};
  \draw[line width=1.5pt, orange!80!black] (1.843909,0.600) -- (2.551016,0.600);
  \fill[orange!80!black] (1.843909,0.600) circle (1.5pt);
  \fill[orange!80!black] (2.551016,0.600) circle (1.5pt);
  \node[left, font=\scriptsize] at (0,1.200) {$(r,v_3)$};
  \draw[line width=1.5pt, orange!80!black] (0.000000,1.200) -- (1.621954,1.200);
  \fill[orange!80!black] (0.000000,1.200) circle (1.5pt);
  \fill[orange!80!black] (1.621954,1.200) circle (1.5pt);
  \node[left, font=\scriptsize] at (0,1.800) {$(v_1,v_2)$};
  \draw[line width=1.5pt, orange!80!black] (0.700000,1.800) -- (1.283095,1.800);
  \fill[orange!80!black] (0.700000,1.800) circle (1.5pt);
  \fill[orange!80!black] (1.283095,1.800) circle (1.5pt);
\end{tikzpicture}}%
    \hspace{0.7cm}
    \resizebox{0.23\linewidth}{!}{\begin{tikzpicture}[x=1.2cm, y=1.2cm]
  \draw[->, thick] (0,0) -- (3.462,0);
  \draw[->, thick] (0,0) -- (0,3.462);
  \node[below, font=\large] at (1.731,-0.600) {death};
  \node[left, font=\large, rotate=90] at (-0.600,1.731) {birth};

  \draw[thick] (0,0) -- (0,-0.08);
  \node[below, font=\large] at (0,-0.08) {0};
  \draw[thick] (0,0) -- (-0.08,0);
  \node[left, font=\large] at (-0.08,0) {0};
  \draw[thick] (1,0) -- (1,-0.08);
  \node[below, font=\large] at (1,-0.08) {1};
  \draw[thick] (0,1) -- (-0.08,1);
  \node[left, font=\large] at (-0.08,1) {1};
  \draw[thick] (2,0) -- (2,-0.08);
  \node[below, font=\large] at (2,-0.08) {2};
  \draw[thick] (0,2) -- (-0.08,2);
  \node[left, font=\large] at (-0.08,2) {2};
  \draw[thick] (3,0) -- (3,-0.08);
  \node[below, font=\large] at (3,-0.08) {3};
  \draw[thick] (0,3) -- (-0.08,3);
  \node[left, font=\large] at (-0.08,3) {3};

  \draw[dashed, black!40] (0,0) -- (3.462,3.462);

  \fill[red!80!black] (0.000000,2.961943) circle (3pt);
  \node[anchor=south west, font=\large] at (0.000000,2.961943) {$(r,v_5)$};
  \fill[red!80!black] (1.843909,2.551016) circle (3pt);
  \node[anchor=south west, font=\large] at (1.843909,2.551016) {$(v_4,v_6)$};
  \fill[red!80!black] (0.000000,1.621954) circle (3pt);
  \node[anchor=south west, font=\large] at (0.000000,1.621954) {$(r,v_3)$};
  \fill[red!80!black] (0.700000,1.283095) circle (3pt);
  \node[anchor=south west, font=\large] at (0.700000,1.283095) {$(v_1,v_2)$};
\end{tikzpicture}}%
    \caption{\textbf{(Left)} A tree with root $r$. The nodes are embedded in $\mathbb{R}^2$. Their coordinates induce Euclidean edge lengths, which define a distance-to-root filtration. \textbf{(Middle)} A barcode showing the resulting $H_0$ persistence under this filtration, with the elder rule deciding which branch survives each merge. The bars are annotated by ($v_d, v_b$), the death and birth nodes respectively. Note that the barcode itself is unlabelled. \textbf{(Right)} The corresponding persistence diagram. Points close to the diagonal represent short-lived features, while points further away represent more persistent features.}
    \label{fig:ph-visualisation}
\end{figure}
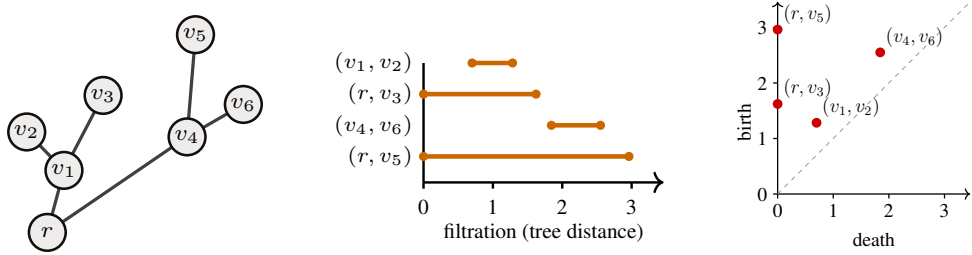

To generate a tree with a \emph{specified} structural profile, we additionally condition on a compact topological descriptor of global morphology \citep{edelsbrunner2008persistent}. This makes the model a controllable source of morphologies: given a descriptor computed from a real cell, it can produce further samples sharing that cell's branching profile -- a form of targeted augmentation an unconditional sampler cannot provide. We do not evaluate downstream augmentation here; our experiments assess only how closely a generated morphology matches the profile it was conditioned on. Below we describe the relevant topological background in the context of tree morphologies.

\paragraph{Filtrations.} Let $T=(V,E,P)$ be a tree with root $r$. Each node $v\in V$ has coordinates $P(v)~\in~\R^3$. A \textit{filtration function} is a map $f:V\to\R$. Given $f$, we define the \textit{superlevel set filtration} of $T$ as the nested sequence of induced subgraphs $\{T_\alpha\}_{\alpha\in\R}$ where $T_\alpha$ contains all nodes $v$ with $f(v)\geq \alpha$. Both filtration functions we use grow away from the root, so sweeping $\alpha$ downwards grows the tree inwards from the tips. 

\paragraph{Barcode and elder rule.} In this work we only use 0-dimensional persistent homology (PH$_0$), which tracks how connected components appear and merge along the filtration. A component is born at the filtration value $\alpha$ when a new connected component appears in $T_\alpha$. Two components merge at filtration value $\alpha$ when they become connected in $T_\alpha$. When two components merge, we apply the \textit{elder rule}: the older component -- the one born at the larger filtration value -- persists, and the younger component dies. The multiset of intervals with endpoints $b_i$ and $d_i$, the birth and death values of component $i$, is called the \textit{persistence barcode} of $T$ under filtration~$f$. See Figure~\ref{fig:ph-visualisation} for an example.

\paragraph{Persistence diagram.} A barcode can equivalently be represented as a multiset of points in $\R^2$, called the \textit{persistence diagram}, where each point $(b_i, d_i)$ corresponds to one bar of the barcode. Points close to the diagonal represent short-lived features, while points further away represent more persistent features. Since $b_i > d_i$ under a superlevel filtration, we follow the convention of plotting death on the horizontal axis, placing the points above the diagonal. 

\paragraph{Filtration choices used in this work.} We use two filtrations, both anchored at the root and both computed with the TMD construction of \citet{kanari2018topological}:
\begin{itemize}
    \item \emph{Path from root} (see Figure~\ref{fig:ph-visualisation}): the geodesic distance from the root along the tree, with each edge weighted by its Euclidean length. This captures the branching structure, yielding a compact summary of the global branching allocation across depth -- \emph{where} and \emph{how much} branching occurs.
    \item \emph{Radial from root}: the straight-line distance $\|P(v)-P(r)\|_2$ from the root. Where the path filtration measures reach \emph{along} the tree, this measures reach \emph{through space}, so the pair distinguishes trees that branch at the same topological depth but at different spatial extents.
\end{itemize}
Both filtrations are invariant to rotation about the root and therefore introduce no dependence on the distinguished axis of the SO(2)-EGNN (Appendix~\ref{appendix: SO(2)-EGNN}). Filtration values are min-max normalised to $[0,1]$ before the barcode is computed, so the descriptor encodes the \emph{shape} of the branching profile rather than absolute scale.

\paragraph{Vectorisation.} To provide this morphological information to the network with a fixed dimension, we convert persistence diagrams into persistence images. A \textit{persistence image} (PI) \citep{adams2017persistence} is a fixed-dimensional vector representation of a persistence diagram obtained by placing Gaussian kernels at each point and discretising the resulting density on a grid. We use a $16\times16$ grid per filtration, so the PIs for the two filtrations flatten and concatenate into a single $512$-dimensional conditioning vector, which is embedded and broadcast to nodes as a graph-level condition. Conditioning on this descriptor produces trees whose branching structure and spatial extent follow the target profile.

\section{Metrics}\label{app:metrics}

This section defines the metrics used in the results tables. We use $T = (V,E,P)$. Let $(x_v,y_v,z_v) = P(v)$ denote the position of a node $v\in V$, let \(r\) be the root, and \(\hat{\mathbf u}\) the unit vector along the chosen principal axis (`up'). 
For nodes $u,v\in V$, let $\boldsymbol{\delta}_{uv}=P(v)-P(u)$ denote the displacement from $u$ to $v$, and let $d_{uv}=\lVert\boldsymbol{\delta}_{uv}\rVert_2$ denote its Euclidean length.
Let \(\mathcal{B}\) denote the set of nodes with at least two children, and \(C(u)\) the children of \(u\). 
We use the following tree statistics:

\metricdef{Axial Extent} Extent of the tree along the principal axis (`height' for botanical trees):
\[         
    \operatorname{AxialExtent}(T)
    = \max_{v\in V}\bigl(P(v)\cdot\hat{\mathbf u}\bigr)
    - \min_{v\in V}\bigl(P(v)\cdot\hat{\mathbf u}\bigr)
\]
\metricdef{Radial Span} Project the tree into the XY-plane, then take the longest distance between any two points in the projection
\[
\operatorname{Diameter}(T) = \max_{u,v \in V}\,
    \bigl\|\,(x_u,y_u)-(x_v,y_v)\,\bigr\|_{2}
\]                                            
\metricdef{Maximum Branch Order} Maximum length of a path from the root to a leaf (in hops):
\[                           
    \operatorname{MaxBranchOrder}(T)
    = \max_{v\in V}\operatorname{len}(r,v),                                                                     
\]
where \(\operatorname{len}(u,v)\) is the distance in hops between $u$ and $v$.
\metricdef{Maximum Path Length} Maximum distance from the root to any node along the edges:
\[
    \operatorname{MaxPathLength}(T)
    = \max_{v\in V}
    \sum_{(u,w)\in\operatorname{path}(r,v)}
    d_{uw}.
\]
\metricdef{Total Edge Length} Sum of the Euclidean lengths of all edges:
\[
    \operatorname{TotalEdgeLength}(T)
    = \sum_{(u,v)\in E}d_{uv}.
\]
\metricdef{Partition Asymmetry} For every inner node $u\in \mathcal B$ we compute its asymmetry as $\frac{|a-b|}{a+b-2}$ where $a$ and $b$ are the number of leaves in the subtree rooted at each of the children of $u$.
We subtract 2 in the denominator since those leaves (one on each side) are unavoidable for any inner node by definition.
Averaging over all $u\in\mathcal B$ gives the partition asymmetry for $T$.                                  
\metricdef{Mean Branch Length} Mean Euclidean length of all edges:
\[
    \operatorname{MeanBranchLength}(T)         
    = \frac{1}{|E|}
    \sum_{(u,v)\in E}d_{uv}
\] 
\metricdef{Critical-Branch Length} Let
$
    V_{\mathrm{crit}}
    = \{r\}\cup\{v\in V:\lvert C(v)\rvert\neq 1\}
$
contain the root, branching nodes, and leaves. A critical branch
$b=(v_0,\ldots,v_k)$ is a maximal path whose endpoints are in
$V_{\mathrm{crit}}$ and whose internal nodes are not. Its length is
\[
    \operatorname{BranchLength}(b)
    = \sum_{i=1}^{k}d_{v_{i-1},v_i}.
\]
\metricdef{Mean Bifurcation Angle} Mean angle, in degrees, across sibling-branch pairs at all branching nodes:
\[
    \operatorname{MeanBifurcationAngle}(T)
    = \frac{1}{|\mathcal{B}|}
    \sum_{u\in\mathcal{B}}
    \frac{180}{\pi}
    \arccos\!\left(
        \frac{
            \boldsymbol{\delta}_{u,c_1(u)}\cdot
            \boldsymbol{\delta}_{u,c_2(u)}
        }{
            d_{u,c_1(u)}\,d_{u,c_2(u)}
        }
    \right)
\]
using $C(u) = (c_1(u),c_2(u))$ since we only consider binary trees. For the root which may have more than two children, we average over all pairs of angles between siblings.     
\metricdef{Critical-Branch Bifurcation Angle} For two critical branches $b_i$ and $b_j$ starting at the same node $u$, let $t(b_i)$ and $t(b_j)$ denote their endpoints. Their angle is
\[
    \theta(b_i,b_j)
    = \frac{180}{\pi}\arccos\!\left(
        \frac{\boldsymbol{\delta}_{u,t(b_i)}\cdot
              \boldsymbol{\delta}_{u,t(b_j)}}
             {d_{u,t(b_i)}\,d_{u,t(b_j)}}
    \right).
\]
\metricdef{Mean Radial Distance to Root} Mean straight-line distance from non-root nodes to the root:
\[                                                 
    \operatorname{MeanRadialToRoot}(T)                   
    = \frac{1}{|V|-1}
    \sum_{v\in V\setminus\{r\}}
    d_{rv}
\]                                                         
\metricdef{Root-to-Leaf Contraction Ratio} Let $\mathcal L$ denote the set of non-root leaves. For each $\ell\in\mathcal L$, the contraction ratio is its straight-line distance from the root divided by its root-to-leaf path length:
\[
    \operatorname{Contraction}(\ell;T)
    = \frac{d_{r\ell}}
            {\displaystyle\sum_{(u,v)\in\operatorname{path}(r,\ell)}
            d_{uv}}.
\]
\metricdef{Mean Contraction} Mean ratio of straight-line root-to-leaf distance to path length:
\[                              
    \operatorname{MeanContraction}(T)
    = \frac{1}{|\mathcal{L}|}                  
    \sum_{\ell\in\mathcal{L}}
    \frac{d_{r\ell}}
            {\displaystyle\sum_{(u,v)\in\operatorname{path}(r,\ell)}
            d_{uv}},
\]                                  
where \(\mathcal{L}\) is the set of non-root leaves.
\metricdef{Sholl Critical Radius}
To compute this we take a sphere around the root and count how many edges cross that sphere. We then compute at which radius this count is maximised.
By normalising with the distance between the root and the furthest node in the tree, the radius becomes comparable between different-size trees.

\paragraph{Evaluation Representations} From these statistics, we form two evaluation representations. For joint-distribution comparisons, each tree is represented by the nine-dimensional morphometric vector comprising axial extent, radial span, maximum branch order, partition asymmetry, mean branch length, mean bifurcation angle, mean radial distance to the root, mean contraction, and Sholl critical radius. This representation is used for morphometric MMD, density, and coverage.

For marginal comparisons, we use six selected rotation-invariant statistics: maximum path length, total edge length, critical-branch length, critical-branch bifurcation angle, root-to-leaf contraction ratio, and partition asymmetry. These statistics do not depend on a chosen global axis, allowing comparison with methods whose outputs have no distinguished orientation.

\paragraph{MMD$^2$}
We compare the generated and ground-truth distributions using the squared maximum mean discrepancy (MMD) with a Gaussian RBF kernel:
\begin{align*}
  \operatorname{MMD}^{2}(P,Q)
  &=
  \mathbb{E}_{x,x'\sim P}[k(x,x')]
  + \mathbb{E}_{y,y'\sim Q}[k(y,y')]
  - 2\,\mathbb{E}_{x\sim P,\,y\sim Q}[k(x,y)],\\
  k(x,y)&=\exp\!\left(-\frac{\|x-y\|_2^2}{2\sigma^2}\right).
\end{align*}
It measures how much the within-distribution similarities exceed cross-distribution similarities and becomes zero if the two distributions $P$ and $Q$ match.

We estimate MMD$^2$ using its unbiased finite-sample estimator, $\widehat{\operatorname{MMD}}_u^2$. Let $G$, $R$, and $R'$ be equally sized sets, where $G$ contains generated morphologies and $R$ and $R'$ are disjoint samples of real training morphologies. We report the baseline-adjusted discrepancy
\[
  \Delta\operatorname{MMD}^{2}
  =
  \widehat{\operatorname{MMD}}_u^{2}(G,R)
  -
  \widehat{\operatorname{MMD}}_u^{2}(R',R).
\]
Values close to zero indicate agreement at the level observed between real samples; small negative values can arise from finite-sample estimation.

\section{Baseline Post-Processing} \label{app:postproc}

This section describes how raw outputs are processed before scoring. We distinguish a method's own tree construction, which is judged as its output, from repair that we add for comparability. Every added step is neutral or in the baseline's favour.

\paragraph{SemlaFlow.} The raw output is a set of coordinates with pairwise bond logits, and the predicted adjacency is not constrained to be a tree. We keep the largest connected component, take its minimum spanning tree by Euclidean edge length, choose as root the unique maximum-degree node when one exists and otherwise the base of the principal axis, prune non-root nodes with more than two children down to the two largest subtrees, and contract non-root degree-$2$ nodes. This yields rooted, connected, and critical trees that are binary except for the root.

The repair is required for a fair evaluation. Root-path and branch-order statistics assume that every node is reachable from one root, persistence calculations require an acyclic graph, and the reference data contain only critical binary nodes and potentially a soma with more than two children. 
We use those repaired graphs to compute the morphology metrics.
For structural validity we resort to the raw outputs to properly measure in how many cases such postprocessing was necessary.

\paragraph{MorphoGen.} The raw output is a $2{,}048$-point cloud on the unit sphere. Tree construction is MorphoGen's published method and its cost is not subtracted: L1-medial contraction, farthest-point subsampling to $1{,}200$ points, soma detection as a local density peak, greedy outward growth of a spanning tree (linker caps non-root nodes at two children) and pruning of short terminal branches. Scale is assigned to each generated tree by a draw from the training-split scale distribution. The resulting tree, which carries about $32$ points per branch, is then contracted to a critical skeleton by removing degree-2 nodes.

We make four changes to the released MorphoGen code: two align the implementation with the method described in the paper, and two tune hyperparameters for MICrONS.
First, the released soma radius of $5$ is in micrometres but is applied to unit-sphere data, so every density ball contains every point and the root is effectively arbitrary; we use a radius calibrated on real clouds, which reduces the normalised soma-placement error from $0.32$ to $0.06$. 
Second, the released code caps all nodes including the root at two children although the paper explicitly exempts the soma; we allow $K_{\max}=23$. 
The two hyperparameters are the seed-edge weight and the branch-pruning length which are swept jointly on the training split and confirmed on validation; on reference clouds they reduce the mean normalised $W_1$ over seven morphometrics from $2.51$ to $0.61$. We use this optimised configuration when reporting on MorphoGen in the main text.

\section{Additional Quantitative Results}
\label{app:additional-quantitative}

This section collects supplementary analyses of the neuron experiments. We examine class-conditioned population structure and test whether repeated TMD-conditioned samples remain specific to their assigned targets.

\subsection{Variability of TMD-Conditioned Sampling}

\begin{figure}[p]
    \centering
    \includegraphics[width=\textwidth,height=0.82\textheight,keepaspectratio]{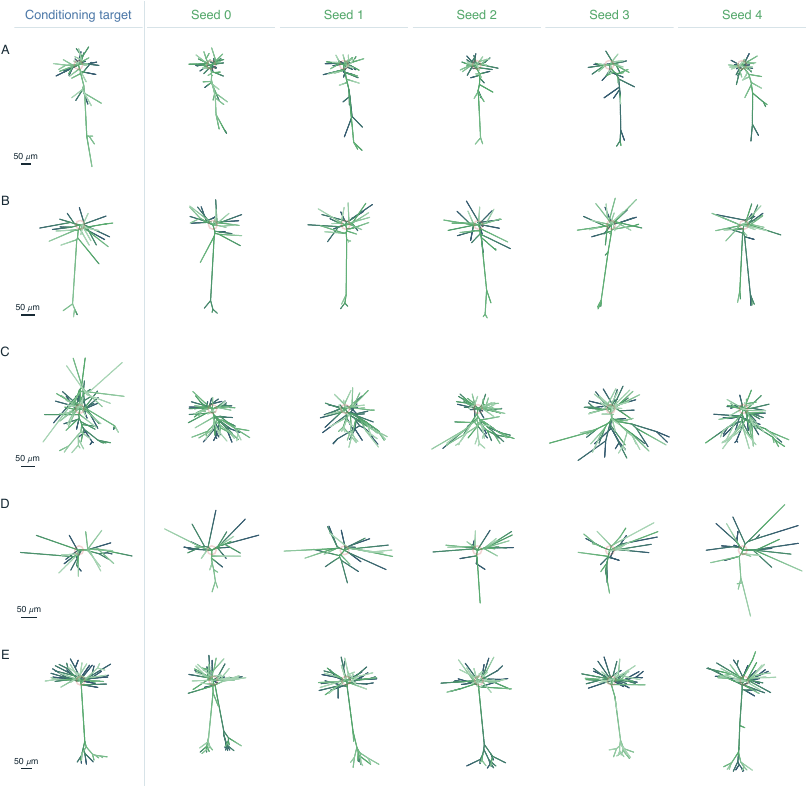}
    \caption{\textbf{Stochastic variation under TMD-conditioning.} Each row shows one held-out conditioning target followed by five independent generations for that target. The generated samples preserve the target’s coarse dendritic organisation while varying in branch geometry and topology.
    Row C shows a more difficult condition for which the generations are visibly more compact than the target. B4 and E0 are unrealistic since they exhibit two apical dendrites instead of just one.}
    \label{fig:appendix_tmd_multiseed_atlas}
\end{figure}

As already indicated in \cref{fig:conditional_samples}, there is some variability in the generation.
Here we quantify this by producing five generations for each of the test neurons with TMD-conditioning.
\cref{tab:tmd-multiseed} shows that five distinct topologies are generated for 97.9\% of targets, and the node count varies for 98.7\%. The persistence-diagram distance among the five variants is half the distance to alternative neurons with the same root degree.
Note that the rows of the table refer to different channels of the persistence diagram (path and radial-root).
Larger sets of repeated samples are shown in \cref{fig:appendix_tmd_multiseed_atlas}.

\begin{table}[ht]
    \centering
    \caption{\textbf{Repeated sampling under TMD-conditioning.} Persistence-diagram distances to the assigned target and to root-degree-matched alternatives across five generations per test neuron. ``assigned closer'' is the percentage of matched comparisons in which the assigned target has the lower distance. Variability between samples is much lower than to arbitrary neurons showing the effect of conditioning.}
    \label{tab:tmd-multiseed}
    \begingroup
\setlength{\tabcolsep}{6pt}
\begin{tabular}{@{}lrrrr@{}}
    \toprule
    \makecell{TMD\\channel} & \makecell{Assigned target\\$W_1$} & \makecell{Matched alternative\\$W_1$} & \makecell{Assigned/matched\\ratio $\downarrow$} & \makecell{Assigned closer\\(\%) $\uparrow$} \\
    \midrule
    Path PD & 2.916 & 5.543 & 0.526 $\pm$ 0.015 & 91.1 $\pm$ 0.8\% \\
    Radial-root PD & 2.775 & 5.666 & 0.490 $\pm$ 0.015 & 90.8 $\pm$ 0.8\% \\
    \bottomrule
\end{tabular}
\endgroup

\end{table}
\FloatBarrier

\subsection{Class-Conditioned Population Structure}

\begin{figure}[!htpb]
    \centering
    \includegraphics[width=0.7\textwidth,keepaspectratio]{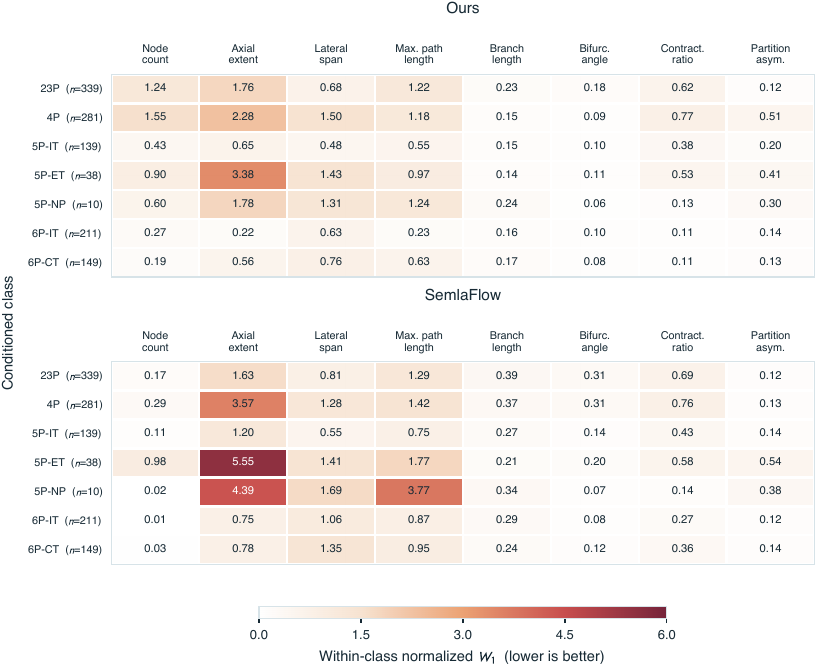}
    \caption{\textbf{Error by cell type.} Class-conditioned generation errors for our model (top) and SemlaFlow (bottom). Entries are class-wise $W_1$ normalised by the reference standard deviation (lower is better). Both models work well on the same cell types and ours clearly outperforms SemlaFlow, except for node count that is given to SemlaFlow as input and partition asymmetry where errors are generally low.}
    \label{fig:appendix_class_morphology}
\end{figure}

In \cref{fig:appendix_class_morphology} we report errors by class for class-conditional generation.
As before, our model receives the soma degree while SemlaFlow gets a reference node count which thus only produces an error when the postprocessing removes nodes after generation, for example when selecting the largest component from a forest.
Our method tends to outperform SemlaFlow over all classes and metrics except for node count as described above and partition asymmetry where the image is mixed.
Also on the axial extent for 23P neurons SemlaFlow is slightly better, while especially on 5P-NP which are hard for SemlaFlow we see a strong advantage of our method. 
Overall, the advantage of our method is visible across cell types and there are clearly easier (5P-IT, 6P-IT) and harder (4P, 5P-ET) cell types for both models.
Also both models tend to work well on the same cell types, except for 5P-NP where SemlaFlow struggles while our method works reasonably well.

\section{Additional Samples} \label{app:additional-samples}

The following galleries expand the examples in the main text, showing unconditional samples across methods, class-conditioned samples, and repeated generations from the same TMD conditions.
Since SemlaFlow and MorphoGen do not have a principal axis, they are plotted such that the root-to-centroid axis is used as such (`up').
Colour encodes signed depth, and translucent red rings mark the roots.

\begin{figure}[p]
    \centering
    \includegraphics[width=\textwidth,height=0.86\textheight,keepaspectratio]{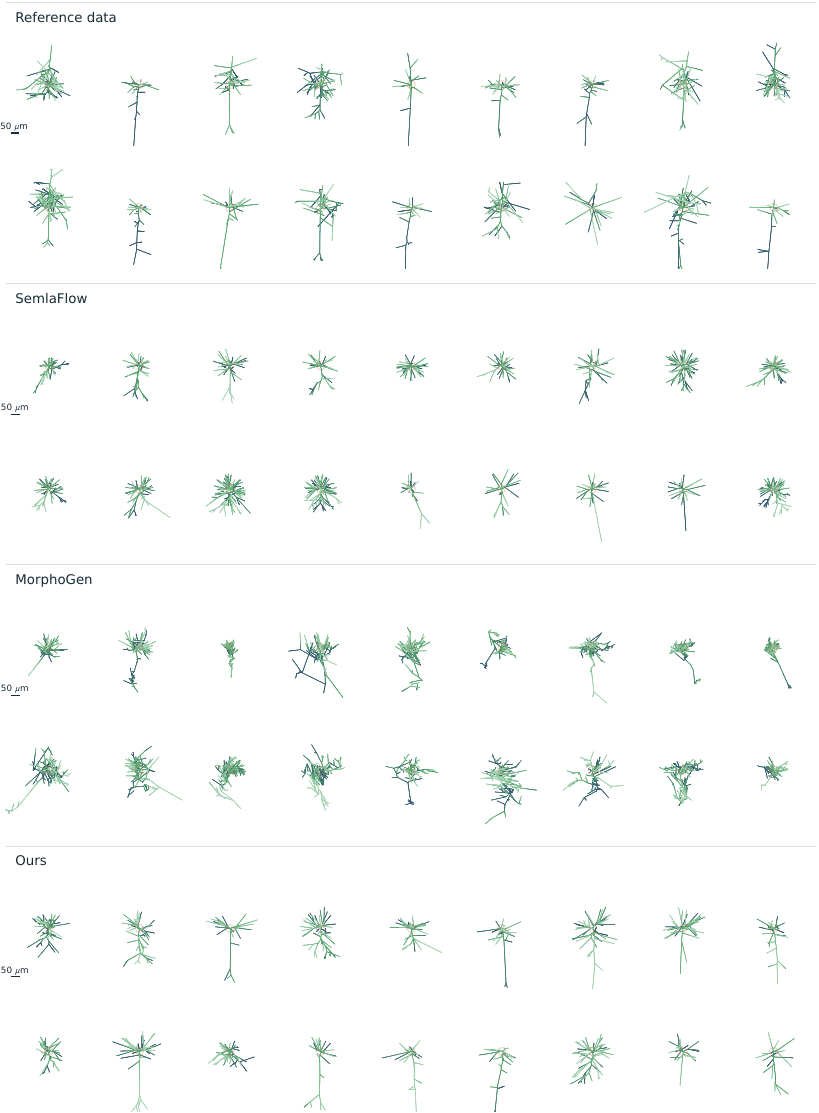}
    \caption{\textbf{Unconditional generation for neurons.} Reference morphologies, post-processed SemlaFlow generations, MorphoGen generations, and generations from our model are shown. Neurons are drawn root-centred and to scale within each method.}
    \label{fig:appendix_unconditional_gallery}
\end{figure}

\begin{figure}[p]
    \centering
    \includegraphics[width=\textwidth,height=0.82\textheight,keepaspectratio]{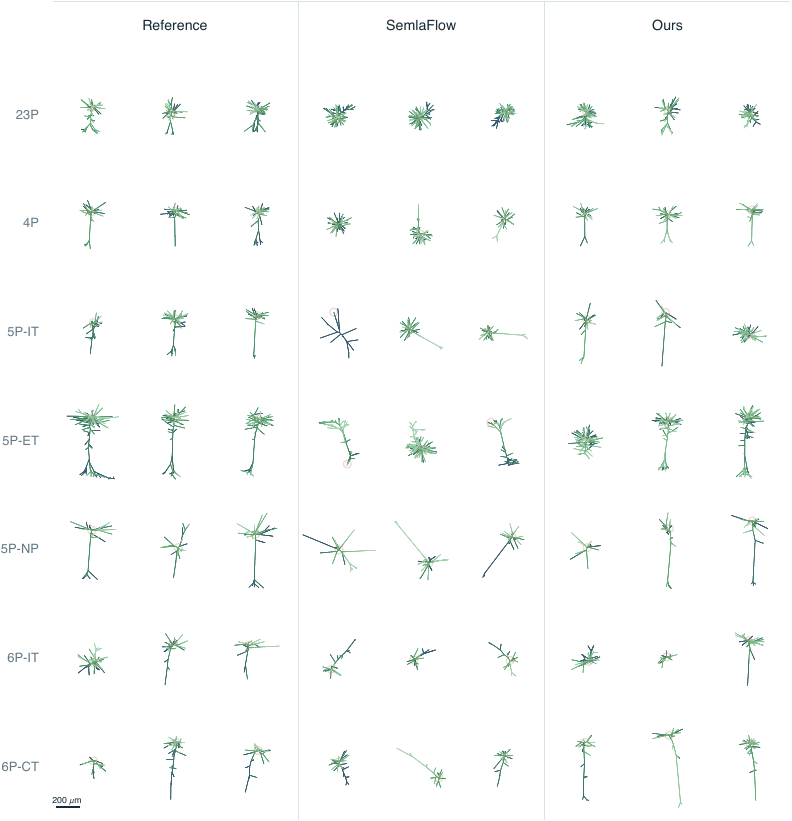}
    \caption{\textbf{Class-conditioned neuron samples.} Each row corresponds to one of the seven cortical pyramidal-cell classes; columns compare training examples, SemlaFlow generations, and generations from our model.
    The displayed samples are selected per source and class rather than treated as paired reconstructions.}
    \label{fig:appendix_class_conditioned_atlas}
\end{figure}

\end{document}